\def\year{2019}\relax
\documentclass[letterpaper]{article} 
\usepackage{aaai19}  
\usepackage{times}  
\usepackage{helvet}  
\usepackage{courier}  
\usepackage{url}  
\usepackage{graphicx}  

\usepackage{array}
\usepackage{hyperref}
\usepackage{etoolbox}
\usepackage{mathtools}
\usepackage{booktabs}
\usepackage{amsmath}
\usepackage{amssymb}
\usepackage{caption}
\usepackage{subcaption,color}

\begin{document}
%
\title{Learning to Compose Topic-Aware Mixture of Experts \\ for Zero-Shot Video Captioning}
\author{Xin Wang$^{\dagger}$, Jiawei Wu$^{\dagger}$, Da Zhang$^{\dagger}$, Yu Su$^{\ddagger}$, William Yang Wang$^{\dagger}$
\\ $^{\dagger}$University of California, Santa Barbara
\\ $^{\ddagger}$The Ohio State University
\\ \tt \{xwang, jiawei\_wu, dazhang, william\}@cs.ucsb.edu, su.809@osu.edu
}
\maketitle

\begin{abstract}
Although promising results have been achieved in video captioning, existing models are limited to the fixed inventory of activities in the training corpus, and do not generalize to open vocabulary scenarios. Here we introduce a novel task, \emph{zero-shot video captioning}, that aims at describing out-of-domain videos of unseen activities. 
Videos of different activities usually require different captioning strategies in many aspects, \textit{i.e.} word selection, semantic construction, and style expression etc, which poses a great challenge to depict novel activities without paired training data. But meanwhile, similar activities share some of those aspects in common. 
Therefore, We propose a principled Topic-Aware Mixture of Experts (TAMoE) model for zero-shot video captioning, which learns to compose different experts based on different topic embeddings, implicitly transferring the knowledge learned from seen activities to unseen ones. Besides, we leverage external topic-related text corpus to construct the topic embedding for each activity, which embodies the most relevant semantic vectors within the topic.
Empirical results not only validate the effectiveness of our method in utilizing semantic knowledge for video captioning, but also show its strong generalization ability when describing novel activities.\footnote{Code is released at \url{ https://github.com/eric-xw/Zero-Shot-Video-Captioning}}
\end{abstract}

\begin{figure}
\begin{center}
\includegraphics[width=0.46\textwidth]{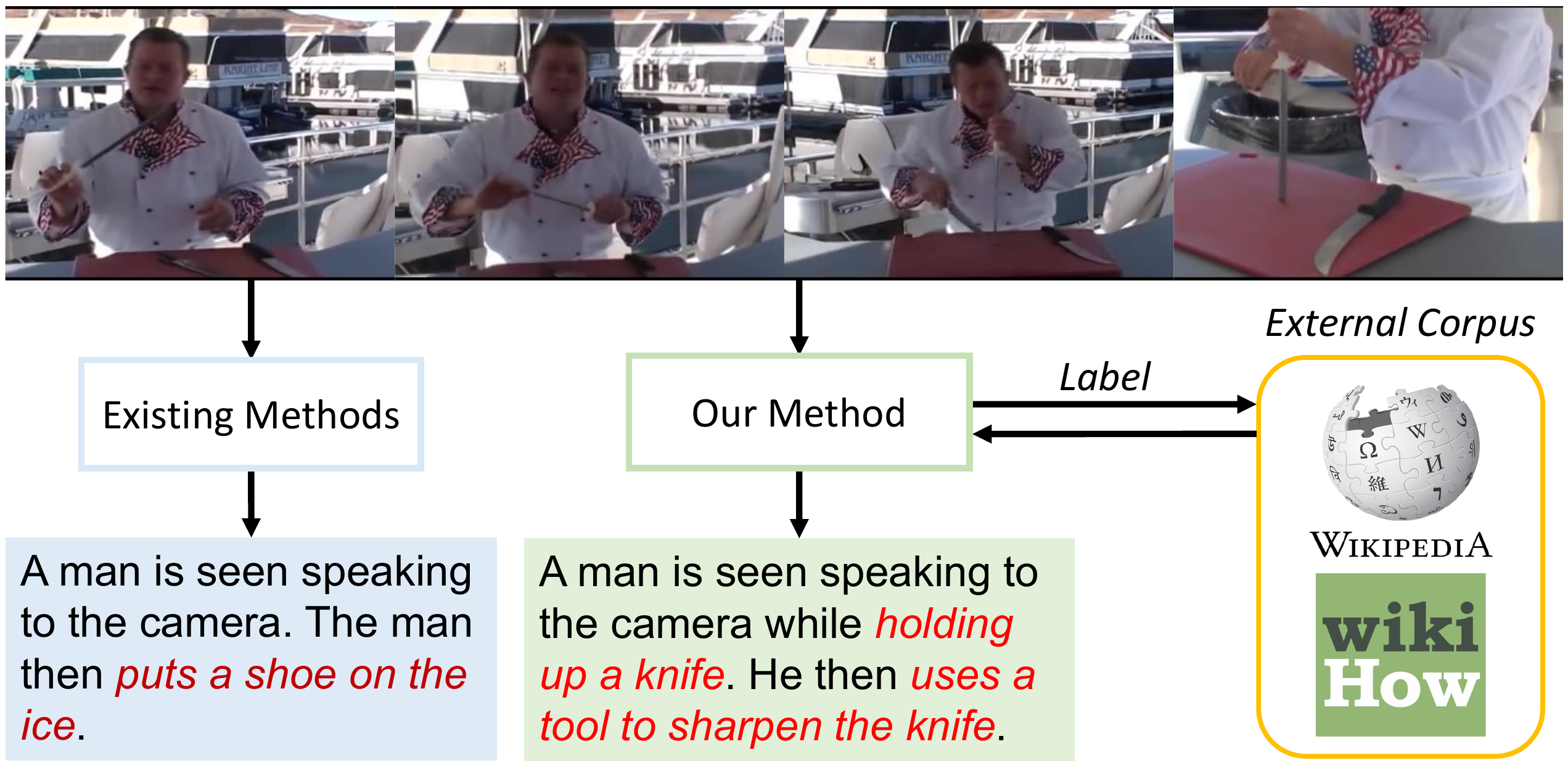}  
\end{center}
   \caption{Example of zero-shot video captioning. ``\emph{Sharpening Knives}'' is a novel activity unseen in training, and existing methods fail to generate a pertinent caption because it is aware of neither the action ``sharpening'' nor the object ``knife". Our method effectively utilizes the knowledge from the external corpus based on predicted activity label and generates a more pertinent caption.}
\label{fig:intro}
\end{figure}

\section{Introduction}

Video captioning aims at automatically describing the content of a video in natural language. It is not only an important testbed for advances in visual understanding and grounded natural language generation, but also has many practical applications such as video search and assisting visually impaired people.
As a result, it has attracted increasing attention in recent years in both NLP~\cite{venugopalan2016improving,wang2018AREL} and computer vision communities~\cite{krishna2017dense}.
Although existing video captioning methods (e.g., sequence-to-sequence model) have achieved promising results, they largely rely on paired videos and textual descriptions for supervision~\cite{xu2016msr}. In other words, they are solely trained to caption the activities that have appeared during training and thus cannot generalize well to novel activities that have never been seen before. However, it is prohibitively expensive to collect paired training data for every possible activity. Therefore, we introduce a new task of \emph{zero-shot video captioning}, where a model is required to accurately describe novel activities in videos without any explicit paired training data. 

An example of zero-shot video captioning is shown in \autoref{fig:intro}, where an existing method fails to correctly caption a video about the novel activity ``\textit{sharpening knives}" because it has learned no knowledge about the activity in training. 
Moreover, the description of different activities vary in word selection, semantic construction, style expression etc, so videos of different activities usually require different captioning strategies, which poses a great challenge in the open vocabulary scenario. Despite the difference, many activities share similar characteristics, \textit{e.g.}, \textit{playing baseball} and \textit{playing football} are both sports activities and a few words can be used to describe both in common. 

Therefore, we propose a novel Topic-Aware Mixture of Experts (TAMoE) approach to caption videos of unseen activities. First, we define a set of \textit{primitive experts} that are sharable by all possible activities, each of which has their own parameters and learns a specialized mapping from latent features to the output vocabulary (the primitive captioning strategies). Then we introduce a \textit{topic-aware gating function} that learns to decide the utilization of those primitive experts and compose a topic-specific captioning model based on a certain topic. Besides, in order to leverage world knowledge from external corpora, we derive a \textit{topic embedding} for each activity from the pretrained semantic embeddings of the most relevant words. When captioning a novel activity, our TAMoE method is capable of inferring the composition of the primitive experts conditioned on the topic embedding, transferring the knowledge learned from seen activities to unseen ones. 
Our main contributions are three-fold:

\begin{itemize}
\item We introduce the task of zero-shot video captioning which aims to accurately describe novel activities in videos without paired training data for the activities.
\item We propose a novel Topic-Aware Mixture of Experts approach for zero-shot video captioning, where a topic-aware gating function learns to infer the utilization of the primitive experts for caption generation from the introduced topic embedding, implicitly doing transfer learning across various topics. 
\item We empirically demonstrate the effectiveness of our method on a popular video captioning dataset and show its strong generalization capability on captioning novel activities.   
\end{itemize}

\begin{figure*}[!t]
\begin{center}
\includegraphics[width=\textwidth]{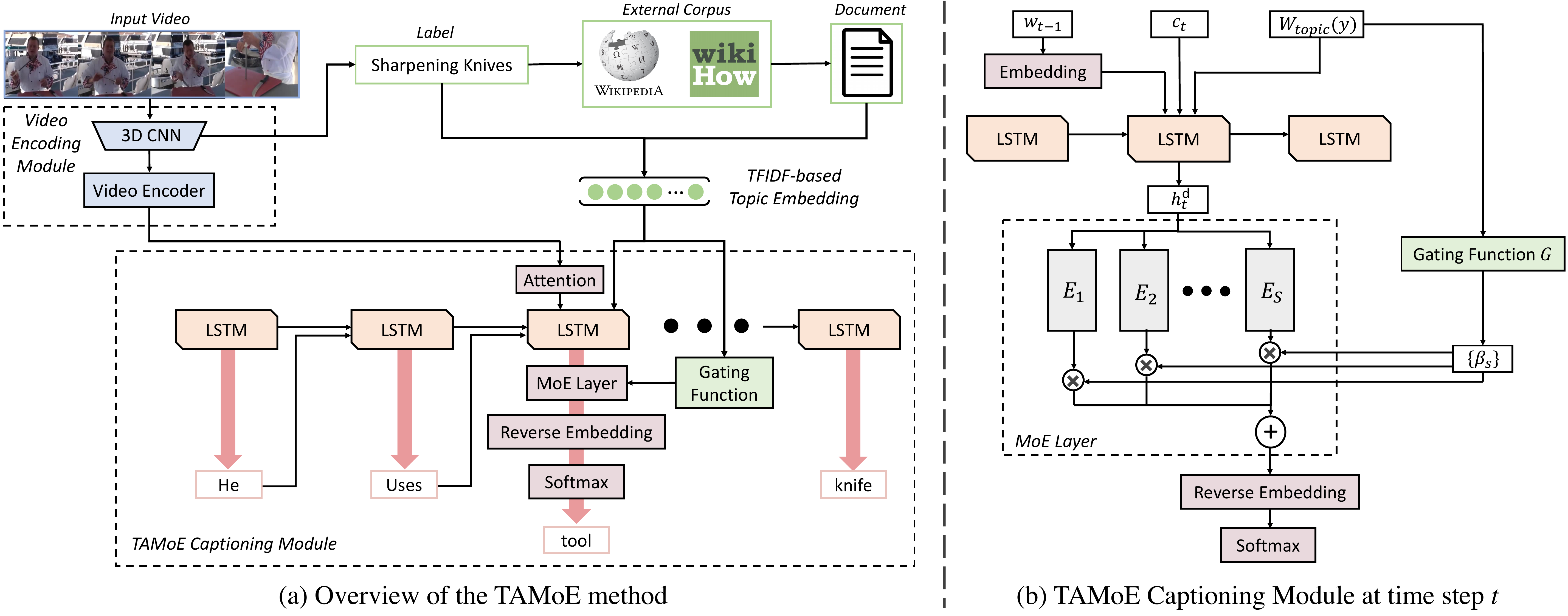}  
\end{center}
\caption{(a) Overview of our TAMoE method; (b) The detailed version of the TAMoE caption module at time step $t$.}
\label{fig:framework}
\end{figure*}

\section{Related Work}
\subsubsection{Video Captioning}
Since S2VT~\cite{venugopalan2015sequence}'s first sequence-to-sequence model for video captioning, numerous improvements have been introduced, such as attention~\cite{yao2015describing}, hierarchical recurrent neural network~\cite{yu2016video,pan2016hierarchical}, multi-modal fusion\cite{gan2016semantic,Shen_2017_CVPR,wang2018watch}, multi-task learning~\cite{pasunuru-bansal:2017:Long}, etc. Meanwhile, a few large-scale datasets are introduced for video captioning, either for single-sentence generation~\cite{xu2016msr} or paragraph generation~\cite{rohrbach14gcpr}. Recently, \cite{krishna2017dense} propose the dense video captioning task, which aims at detecting multiple events that occur in a video and describing each of them. However, existing methods mainly focus on learning from paired training data and testing on similar videos. Though some work has attempted to utilize linguistic knowledge to assist video captioning~\cite{thomason2014integrating,venugopalan2016improving}, none of them has formally considered zero-shot video captioning to describe videos of novel activities, which is the focus of this study.     

\subsubsection{Novel Object Captioning in Images}
Recent studies on novel object captioning~\cite{anne2016deep,venugopalan2016captioning} attempt to describe novel objects not appearing during training. Zero-shot video captioning shares a similar spirit in the sense that it also generates captions without paired data. But zero-shot video captioning is a more challenging task: images are static scenes, and methods based on noun word replacement can perform well on novel object captioning~\cite{anderson-EtAl:2017:EMNLP2017,wu2018decoupled,lu2018neural}; while describing novel activities in videos requires both temporal understanding of videos and deeper understanding of the social or human knowledge of activities beyond the object level. Different activities need different captioning strategies, as well as share some common characteristics. Motivated by this, our method learns the underlying mapping experts from the latent representations to the vocabulary, with a topic-aware gating mechanism implicitly transferring the utilization, which is orthogonal to these methods for novel object captioning in images.

\subsubsection{Zero-Shot Activity Recognition}
In prior work, zero-shot learning has been studied on the task of activity recognition~\cite{caba2015activitynet,zhang2018bmvc}, to predict a previously unseen activity. Unlike zero-shot activity recognition~\cite{gan2015exploring,gan2016concepts,zellers2017zero}, zero-shot video captioning focuses on the language generation part---learning to describe out-of-domain videos of a novel activity without paired captions but with the knowledge of the activity. 
This technique is valuable because caption annotations for videos are much more expensive to get compared with activity labels. 

\subsubsection{Mixture of Experts}
Mixture of Experts (MoE) is originally formulated by \citeauthor{jacobs1991adaptive}, which learns to compose multiple expert networks with each to handle a subset of the training cases. Then MoE has been applied to various machine learning algorithms~\cite{jordan1994hierarchical,collobert2003scaling}, such as SVMs~\cite{collobert2002parallel}, Gaussian Processes~\cite{tresp2001mixtures}, and deep networks~\cite{ahmed2016network,wang2018deep,gu2018naacl}.
Recently, \citeauthor{shazeer2017outrageously} proposes a sparsely-gated mixture-of-experts layer for language modeling, which benefits from conditional computation. \citeauthor{yang2018breaking} extends it to Mixture of Softmax to break the softmax bottleneck and thus increase the capacity of the language model.
In this work, we exploit the nature of MoE for transfer learning by training a topic-aware gating function to compose primitive experts and adapt to various topics.

\section{Describing Novel Activities in Videos}
\subsection{Task Definition}
Here we first introduce the general video captioning task, whose input is a sequence of video frames $\nu = \{v_1, v_2, \ldots, v_n\}$ where $n$ is the number of frames in temporal order. The output is a sequence of words $\mathcal{W} = \{w_1, w_2, \ldots, w_T\}$, where $T$ is the length of the generated word sequence. At each time step $t$, a model chooses a word $w_t$ from a vocabulary $V$ that is built from the paired training corpus. Normally, the vocabulary $V$ can cover the possible output tokens if tested on the same activities as in training. But for zero-shot video captioning, the testing videos are about novel activities that have never been seen during training and require many out-of-vocabulary words to describe. So zero-shot video captioning is an open vocabulary learning scenario, whose objective is to produce a word sequence $\mathcal{W}$ with $w_t \in V^*$, where $V^*$ is beyond the training corpus and ideally would consist of all the possible tokens from the world knowledge. But in practice, we narrow it down to the vocabulary related to all the activities in the dataset.

\subsection{Method Overview}
We show in \autoref{fig:framework}(a) the overall pipeline of our Topic-Aware Mixture of Experts (TAMoE) approach, which mainly consists of the \textit{video encoding module}, the \textit{TFIDF-based topic embedding}, and the \textit{TAMoE captioning module}. The \textit{video encoding module} encodes video-level features and predicts the activity label. Then, the topic-related documents can be fetched from the external corpus and used to calculate the \textit{TFIDF-based topic embedding}, which represents the semantic meaning of the activity. In the decoding stage, the \textit{TAMoE captioning module} takes both the video features and the topic embedding as input and generates the caption by dynamically composing specialized experts.
In the following sections, we discuss each module in details. 

\subsection{Video Encoding Module} 
\label{sec:video}
Given the input video $\nu = \{v_1, v_2, \ldots, v_n\}$, we employ the pretrained 3D convolutional neural networks to extract the segment-level features $\{f_j\}$ where $j=1,2,\ldots,m \ll n$ (we use I3D features in our experiments\footnote{I3D~\cite{carreira2017quo} is the state-of-the-art 3D CNN model for video classification.}). The I3D features include short-range temporal dynamics while keeping advanced spatial representations. Then our model sends the segment-level features $\{f_j\}$ to the video encoder, which is a bidirectional LSTM, to model long-range temporal contexts. It outputs the hidden representations $\{h^e_j\}$ with $e$ denoting the video encoder, which encodes the video-level features.

\subsection{TFIDF-based Topic Embedding}
To learn the knowledge of the activities without paired captions, we fetch topic-related documents from various data sources, e.g., Wikipedia and WikiHow. We also employ the pretrained \textit{fasttext} embeddings~\cite{mikolov2018advances} to calculate the representations of the topics\footnote{Though we use fasttext embeddings here, our method is not limited to a particular word embeddings.} Given an activity label $y$ ($y \in Y$) and the related documents $D_y$, we need to compute the topic-specific knowledge representations. 

The documents contain many high-frequency but irrelevant words, e.g., \textit{the, to, a}, so average embedding is too noisy to effectively represent the knowledge of the topic. Term Frequency-Inverse Document Frequency (TF-IDF) is an efficient statistical method to reflect the importance of a word to a document. Here we propose a topic-aware TF-IDF weighting $g_k(y)$ to calculate the relevance of each unigram $x_k$ to the topic-related documents $D_y$:
\begin{equation}
g_k(y) = \frac{z_k(y)}{\sum_{x_l \in D_y} z_l(y)} \log (\frac{|Y|}{\sum_{y' \in Y} \min(1, z_k(y'))})
\end{equation}
where $z_k(y)$ is the number of times the unigram $x_k$ occurs in the documents $D_y$ related to label $y$. The first term is the term frequency of the unigram $x_k$, which places a higher weight on words that frequently occur in the topic-related documents $D_y$. The second term measures the rarity of $x_k$ with inverse document frequency, reducing the weight if $x_k$ commonly exists across all the topics.
Then our TF-IDF embedding is
\begin{equation}\label{eq:tfidf}
W_{tfidf}(y) = \sum_{x_k \in D_y} g_k(y) W_{fasttext}(x_k) 
\end{equation}
where $W_{fasttext}$ denotes the pretrained fasttext embeddings. 
As shown in \autoref{fig:framework}(a), the TF-IDF embedding is concatenated with the average embedding of the activity label and eventually taken as the topic embedding $W_{topic}(y)$. 

\subsection{TAMoE Captioning Module}

\paragraph{Attention-based Decoder LSTM}
The backbone of the captioning model is an attention-based LSTM.
At each time step $t$ in the decoding stage, the decoder LSTM produces its output $h_t^d$ ($d$ denoting the decoder) by considering the word at previous step $w_{t-1}$, the visual context vector $c_t$, the topic embedding $W_{topic}(y)$ and its internal hidden state $h_{t-1}^d$. In formula, 
\begin{equation}\label{eq:de_lstm}
h_t^d = LSTM([w_{t-1}, c_t, W_{topic}(y)], h_{t-1}^d)
\end{equation}
where the context vector $c_t$ is a weighted sum of the encoded video features $\{h_j^e\}$
\begin{equation}
c_t = \sum \alpha_{t,j} h_j^e
\end{equation}
These attention weights $\{\alpha_{t,j}\}$ act as an alignment mechanism by giving higher weights to certain features that allow better prediction. They are learned by the attention mechanism proposed in \cite{bahdanau2014neural}.

\paragraph{Mixture-of-Expert Layer and Topic-Aware Gating Function}
Following \autoref{eq:de_lstm}, the output of the decoder LSTM $h_t^d$ is then fed into the Mixture-of-Experts (MoE) layer (see \autoref{fig:framework}(b)). Here each expert is an underlying mapping function from the latent representation $h_t^d$ to the vocabulary, which learns the captioning primitives that are shareable to all topics. All the experts in the same MoE layer have the same architecture, which is parameterized by a fully-connected layer and a nonlinear ReLU activation.   
Let $S$ denote the number of experts and $E_s$ be the $s$-th expert, then output of the MoE layer is 
\begin{equation}\label{eq:moe}
o_t = \sum_{s=1}^S \beta_s E_s(h_t^d)
\end{equation}
where $\beta_s$ is the gating weight of the expert $E_s$, representing the utilization of the expert $E_s$. And it is determined by the topic-aware gating function $G$: 
\begin{equation}
\beta_s = \frac{\exp(G(W_{topic}(y))_s / \tau)}{\sum_{i=1}^S \exp(G(W_{topic}(y))_i / \tau)}
\end{equation}
where $G$ is a multilayer perceptron in our model. The temperature $\tau$ determines the diversity of the gating weights. 
The topic-aware gating function $G$ is conditioned on the topic embedding $W_{topic}(y)$ and learns to combine the expertise of those primitive experts for a certain topic. Intuitively, $G$ learns topic-aware language dynamics and composes different expert utilization for different topics based on the topic embeddings, which can implicitly transfer the utilization across topics.

\paragraph{Embedding and Reverse Embedding Layers}
In addition, we also employ semantic word embeddings in our captioning model to help generate descriptions of unseen activities. 
Incorporating pretrained embeddings assigns semantic meanings to those out-of-domain words and thus can facilitate the open vocabulary learning~\cite{venugopalan2016captioning}. Particularly, we load the fasttext embeddings into both the embedding layer and the reverse embedding layer (see \autoref{fig:framework}(b)), and freeze their weights during training. So the embedding layer represents the input word (one-hot vector) into semantically meaningful dense vectors, while the reverse embedding layer is placed before the softmax layer to reverse the mapping from the feature vectors into the vocabulary space. 

\subsection{Learning}
\paragraph{Cross Entropy Loss}
We adopt the cross entropy loss to train our models.
Let $\theta$ denote the model parameters and $w_{1:T}^*$ be the ground-truth word sequence, then the training loss is defined as
\begin{equation}\label{eq:loss}
\mathcal{L}(\theta) = - \sum_{t=1}^T \log p(w_t^* | w_{1:t-1}^*, \theta)
\end{equation}
where is
where $p(w_t|w_{1:t-1},\theta)$ is the probability distribution of the next word.

\paragraph{Variational Dropout}
In order to regularize our MoE layer and promote expert diversity, we adopt the variational dropout~\cite{gal2016theoretically,merity2017regularizing} when training the TAMoE module.  
Different from the standard dropout, variational dropout samples a binary dropout mask only once upon the first call and then repeatedly uses that locked dropout mask within samples.
In addition, the variational dropout helps stabilize the training of the topic-aware gating mechanism by making the expert behaviors consistent within samples. 

\begin{table*}[t]
\small
\renewcommand{\arraystretch}{1.1}
\setlength{\tabcolsep}{4pt}
\begin{center}
  \begin{tabular}{ l l | cccccccc | ccccccc }
      \toprule
      &  &  \multicolumn{7}{c}{\textbf{Seen Test Set}} & \, &\multicolumn{7}{c}{\textbf{Unseen Test Set}} \\
      \midrule
           
        Model & Embedding
        & CIDEr & B-1 & B-2 & B-3 & B-4 & M & R \, &
        & CIDEr & B-1 & B-2 & B-3 & B-4 & M & R \\
        \midrule
        
        Base & task-specific & 
        29.67 & 23.57 & 12.06 & 7.02 & 4.42 & 9.77 & 21.45 & \, &
        21.59 & 22.34 & 10.57 & 5.76 & 3.45 & 9.01 & 20.06\\
        Base & fasttext & 
        31.48 & 23.88 & 12.20 & 7.11 & 4.39 & 10.16 & 21.69 & \, &
        22.51 & 22.50 & 11.01 & 6.02 & 3.58 & 9.43 & 20.70\\
        \midrule
        Topic & task-specific & 
        33.06 & 24.48 & 12.64 & 7.27 & 4.32 & 10.49 & 22.24 & \, &
        23.06 & 22.06 & 10.34 & 6.05 & 3.62 & 9.40 & 20.60 \\
        Topic & fasttext & 
        33.72 & 24.53 & 12.56 & 7.20 & 4.44 & 10.24 & 22.11 & \, &
        24.06 & 22.97 & 11.09 & 5.98 & 3.51 & 9.70 & 20.98 \\
        \midrule
        TAMoE  & task-specific &
        34.38 & \textbf{25.79} & 13.29 & \textbf{7.44} & 4.46 & 10.69 & \textbf{23.03} & \, &
        24.39 & 23.36 & \textbf{11.19} & 6.05 & 3.59 & 9.28 & \textbf{21.46} \\
        TAMoE  & fasttext &
           \textbf{35.53} & 25.51 & \textbf{13.93} & 7.39 & \textbf{4.61} & \textbf{10.83} & 22.51 & \, &
        \textbf{28.23} & \textbf{24.34} & 11.18 & \textbf{6.14} & \textbf{3.68} & \textbf{9.96} & 21.17 \\
        \bottomrule
  \end{tabular}
\end{center}
\caption{
Comparison with the baseline methods on the held-out ActivityNet-Captions dataset. 
We report the results of our TAMoE model and the other baseline models in terms of CIDEr, BLEU (B), METEOR (M), and ROUGE-L (R) scores.
}
\label{table:overall}
\end{table*} 

\section{Experimental Setup}
\subsection{Held-out ActivityNet-Captions Dataset}
ActivityNet~\cite{caba2015activitynet} is a well-known benchmark for video classification and detection, which covers 200 classes of activities. 
Recently, \cite{krishna2017dense} have collected the corresponding natural language description for the videos in the ActivityNet dataset, leading to the ActivityNet-Captions dataset. 
We set up the zero-shot learning scenario based on the ActivityNet-Captions dataset.
We re-split the videos of the 200 activities into the the \textit{training set} (170 activities), the \textit{validation set} (15 activities), and the \textit{unseen test set} (15 activities). Each activity is unique and only exists in one split above. We hold out the novel 15 activities\footnote{The held-out 15 activities are: \textit{``making a lemonade", ``armwrestling", “longboarding”, “playing badminton”, ``shuffleboard", ``slacklining", ``hula hoop", ``playing drums", ``braiding hair", ``gargling mouthwash", ``installing carpet", ``sharpening knives", ``grooming dog", ``assembling bicycle", ``painting fence"}.} for testing that appear during neither training nor validation. 
In order to compare with the model's performance on the supervised split, we then further split an additional \textit{seen test set} that shares the same activities with the training set but has different video samples. 
The external text corpus is crawled from Wikipedia, WikiHow, and some related documents in the first Google Search page. On average there are 2.72 related documents per activity (the max is 10).  

\subsection{Evaluation Metrics}
We use four popular and diverse metrics for language generation, CIDEr, BLEU, METEOR, and ROUGE-L. 
Among these metrics, only CIDEr weighs the topic relevance of n-grams and thus can better reflect a model's capability on captioning novel activities. Therefore, we use CIDEr as the major metric. In addition to the average CIDEr score of the n-grams ($n=1,2,3,4$), we also report individual CIDEr-1, CIDEr-2, CIDEr-3, and CIDEr-4 scores. 

\subsection{Implementation Details}
To preprocess the videos, we sample each video at $20fps$ and extract the I3D features~\cite{carreira2017quo} from these sampled frames. Note that the I3D model is pre-trained on the Kinects dataset~\cite{kay2017kinetics} and used here without fine-tuning. 
The activity labels feeding to our model are predicted by a pretrained 3D CNN model~\cite{wang2016temporal} for activity classification. 
The vocabulary is built based on the training corpus and the unpaired external corpus. We use 300-dimensional pretrained fasttext embedding for words.  
All the hyper-parameters are tuned on the validation set. The maximum number of video features is 200 and the maximum caption length is 32. The video encoder is a biLSTM of size 512, and the decoder LSTM is of size 1024. We initialize all the parameters from a uniform distribution on $[-0.1, 0.1]$. Adadelta optimizer~\cite{zeiler2012adadelta} is used with batch size 64. Learning rate starts at 1 and is then halved when the current CIDEr score does not surpass the previous best in 4 epochs. The maximum number of epochs is 100, and we shuffle the training data at each epoch. Schedule sampling~\cite{bengio2015scheduled} is also employed to train the models. Beam search of size 5 is used at test time. It takes around 6 hours to fully train a model on a TITAN X. 

\section{Experiments and Analysis}

We compare three models on the Held-out ActivityNet-Captions dataset.

\noindent \textbf{Base}: we first implement the state-of-the-art attention-based sequence-to-sequence model used in \cite{wang2018video} as our baseline (\textit{Base}). 
Simply put, the Base model is the model in \autoref{fig:framework} without the topic embedding module and the gating function. Everything else is exactly the same.

\noindent \textbf{Topic}: the Topic model has a very similar architecture with the Base model, except that its decoder takes the proposed topic embedding as an additional input.  

\noindent \textbf{TAMoE}: the proposed TAMoE model is illustrated in \autoref{fig:framework}, which consists of the video encoding module, the topic embedding, the topic-ware gating function, and the Mixture-of-Experts layer.

Moreover, we test the impact of pretrained word embeddings by comparing two word embedding initialization strategies: (1) \textit{task-specific}, that randomly initializes the embeddings and learns them during training, and (2) \textit{fasttext}, that uses pretrained fasttext embeddings (fixed in training).

\subsection{Experimental Results}
\subsubsection{Seen and Unseen Test Sets of the Held-out ActivityNet Captions}
\autoref{table:overall} shows the results on both the seen and the unseen test sets. First, it can be noted that incorporating pretrained fasttext embeddings brings a consistent improvement across models on both test sets, especially for the zero-shot learning scenario on the unseen test set. 
Second, by comparing the Base model and the Topic model it can be observed that solely adding the proposed topic embedding can bring some improvement. 
These validate the hypothesis that the pretrained embeddings can bring useful prior knowledge to assist caption generation, and it facilitates the generation of out-of-domain words that do not appear in the training data. 
More importantly, our TAMoE model significantly improves the scores over the baseline models. For instance, our full TAMoE model outperforms the Base model on both the seen and the unseen test sets by a large margin, with respectively 19.75\% and 30.75\% relative improvement on CIDEr. 
The remarkable improvement on the unseen test set clearly demonstrates the superior capability of the proposed model on captioning novel activities.

Because CIDEr is the only metric that considers the informativeness of the generated captions by penalizing uninformative n-grams that frequently occur across the dataset, it is expected that model performance will present a larger gap on CIDEr between the seen and the unseen test sets. This is confirmed by our results, which reinforces that CIDEr is a better metric for the task of novel activity captioning because it makes a more clear distinction between common n-grams that occur across all activities and activity-specific n-grams.
Therefore, we will use CIDEr hereafter.

\begin{table}[t]
\small
\renewcommand{\arraystretch}{1.1}
\setlength{\tabcolsep}{5pt}
\begin{center}
  \begin{tabular}{ l c c c c }
          \toprule
        Model
           & CIDEr & BLEU-4 & METEOR & ROUGE-L \\
        \midrule
        Base & 47.2 & 40.9 & 28.8 & 60.9 \\
        TAMoE & 48.9 & 42.2 & 29.4 & 62.0 \\
        \bottomrule
  \end{tabular}
\end{center}
\caption{Results on the MSR-VTT dataset.}
\label{table:msrvtt}
\end{table} 

\begin{table}[t]
\small
\renewcommand{\arraystretch}{1.1}
\setlength{\tabcolsep}{5pt}
\begin{center}
  \begin{tabular}{ l l c c c c}
          \toprule
        Model & Embedding
           & C-1 & C-2 & C-3 & C-4 \\
        \midrule

        Base & task-specific & 
        52.13 & 20.41 & 8.92 & 4.18 \\
        Base & fasttext & 
        55.17 & 21.40 & 8.81 & 4.64 \\
        \midrule
        Topic & task-specific & 
        55.79 & 20.94 & 9.47 & 4.68 \\
           Topic & fasttext & 
        58.84 & 23.33 & 9.32 & 4.75 \\
        \midrule
        TAMoE & task-specific &  
        58.81 & 22.98 & 10.42 & 6.00\\
        TAMoE & fasttext & 
        \textbf{67.48} & \textbf{25.89} & \textbf{12.09} & \textbf{7.47}\\
        \bottomrule
  \end{tabular}
\end{center}
\caption{Individual CIDEr scores of unigrams (C-1), bigrams (C-2), trigrams (C-3), and fourgrams (C-4) on the unseen test set, which are all novel activities.}
\label{table:cider}
\end{table} 

\begin{table}[t]
\small
\renewcommand{\arraystretch}{1.1}
\setlength{\tabcolsep}{3pt}
\begin{center}
  \begin{tabular}{ c | c | c | c | c | c}
          \toprule
        I3D Video Features & \checkmark & \checkmark & \checkmark &  & \checkmark  \\
        Average Label Embedding & & \checkmark &  & \checkmark & \checkmark \\
        TFIDF Embedding & & & \checkmark & \checkmark & \checkmark \\
        \midrule
        CIDEr & 22.51 & 25.96 & 26.61 & 15.77 & 28.23 \\
        \bottomrule
  \end{tabular}
\end{center}
\caption{Impact of different features on the TAMoE model. \textit{I3D Video Features} are the extracted video features using the pretrained I3D model; \textit{Average Label Embedding} is the average embedding of the words in the predicted activity label; \textit{TFIDF Embedding} is the weighted embedding of the external topic-related documents (see \autoref{eq:tfidf}).}
\label{table:feature}
\end{table} 

\subsubsection{MSR-VTT} To prove the effectiveness of our method on generic video captioning, we further test it on the widely-used MSR-VTT dataset~\cite{xu2016msr}. As shown in \autoref{table:msrvtt}, the TAMoE approach outperforms the Base model on all the metrics by a large margin. Note that for simplicity, we utilize the pretrained visual and audio features as used in \cite{wang2018watch} as well as the ground-truth category labels on this dataset. 

\subsection{Ablation Study}

\subsubsection{Evaluation on Different N-grams}
In order to take a closer look at the transfer influence of our TAMoE model on individual n-grams, we calculate the CIDEr score of unigrams, bigrams, trigrams, and fourgrams on the unseen test set separately. As seen in \autoref{table:cider}, our TAMoE model performs the best on all n-grams, but the CIDEr score of 4-grams is still not very satisfactory. A general limitation of current captioning systems is that the focus is still on learning word-level embeddings and generating a caption word by word. Incorporating phrase-level embeddings may alleviate this issue. We leave it for future study.    

\subsubsection{Impact of Different Features}
In \autoref{table:feature}, we test the influence of the I3D video features and various versions of the topic embedding. Evidently, it performs the best to use the concatenation of the average label embedding and the TFIDF embedding from external corpus as the topic embedding. Besides, without videos features, the model is unable to generate diverse captions for different videos that also match the video content (the corresponding CIDEr score is as low as 15.77).

\begin{figure}[!t]
\begin{center}
\includegraphics[width=0.45\textwidth]{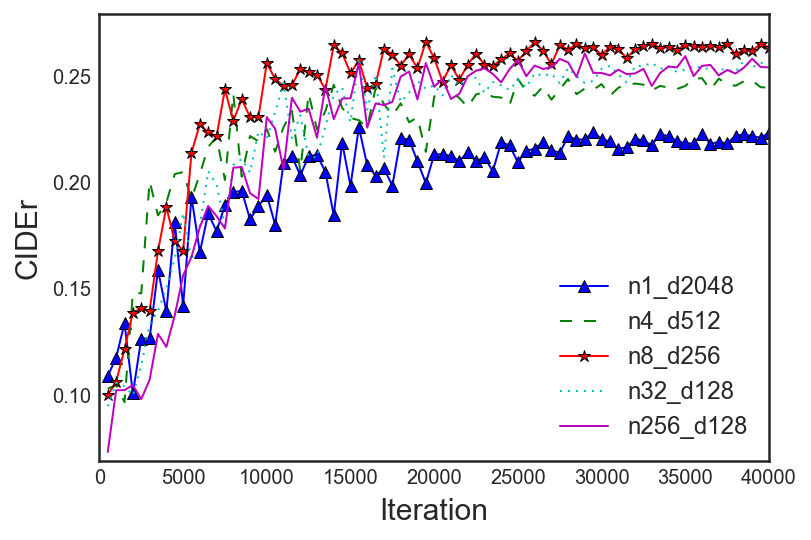}  
\end{center}
   \caption{Learning curves of the TAMoE models with different numbers of experts (\textit{n}) and different expert dimension (\textit{d}). For example, \textit{n4\_d512} denotes the TAMoE model with 4 experts, each of dimension 512. Note the validation scores are calculated by greedy decoding, which are lower than than testing scores by beam search of size 5.}
\label{fig:curve}
\end{figure}

\begin{figure*}
\begin{center}
\includegraphics[width=\textwidth]{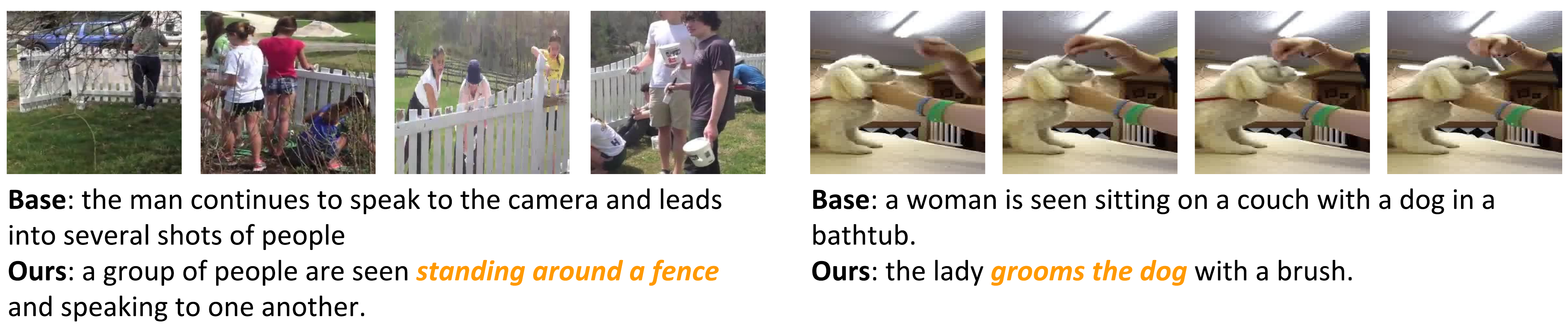}  
\end{center}
\vspace{-2ex}
   \caption{Qualitative comparison between our TAMoE model and the Base model on describing novel activities.}
\label{fig:demo}
\end{figure*}

\begin{table}
\small
\renewcommand{\arraystretch}{1.1}
\setlength{\tabcolsep}{2pt}
\begin{center}
  \scalebox{0.88}{
  \begin{tabular}{ l l l l }           
        \toprule
        \textbf{Novel Activity} & \textbf{Base} & \textbf{TAMoE} & \textbf{Top-4 related words} \\
        \midrule
        making a lemonnade 
        & 28.63 & 31.66
        & lemonade, sugar, lemon, juice\\
        arm wrestling 
        & 23.72 & 35.96
        & wrestling, arm, opponent, strength\\
        longboarding 
        & 20.51 & 28.79
        & longboard, board, foot, riding\\
        playing badminton 
        & 20.18 & 22.00
        & shuttle, racket, shuttlecock, court\\
        shuffleboard 
        & 14.95 & 20.85
        & shuffleboard, disks, discs, puck\\
        slacklining 
        & 24.43 & 21.33
        & slackline, slacklining, line, balance\\
        hula hoop 
        & 17.50 & 26.29
        & hoop, hula, hoops, waist\\
        playing drums 
        & 31.70 & 39.44
        & drum, snare, metronome, hat\\
        braiding hair 
        & 21.30 & 36.80
        & braid, hair, section, strands\\
        gargling mouthwash
        & 11.09 & 52.03
        & mouthwash, mouth, gargling, fluoride\\
        installing carpet 
        & 22.40 & 17.85
        & carpet, strips, tackless, wall\\
        sharpening knives 
        & 24.77 & 43.63
        & stone, knife, sharpening, blade\\
        grooming dog 
        & 18.33 & 26.61
        & dog, clippers, shampoo, fur\\
        assembling bicycle 
        & 22.17 & 28.74
        & handlebar, bike, stem, seat\\
        painting fence 
        & 23.56 & 23.15
        & fence, paint, painting, sprayer\\
        \bottomrule
  \end{tabular}}
\end{center}
\caption{Topic-wise comparison. We compare the CIDEr scores of the Base model and our TAMoE model within each activity. In the right-most column, we list the top words based on their TF-IDF weights in the external topic-related documents.}
\label{table:pertopic}
\end{table}    

\subsubsection{Impact of The Number of Experts}
An important hyper-parameter in our TAMoE model is the number of experts in the Mixture-of-Experts layer. We compare models with different numbers of experts. For a fair comparison, we adjust the dimensionality of each expert to ensure that different models have the same capacity (number of parameters). Note that we set the minimum expert dimensionality as 128 to ensure a lower bound of each expert's capacity. Their learning curves on the validation set are shown in \autoref{fig:curve}. As can be observed, the model with 8 experts of dimension 256 (\textit{n8\_d256}) works the best, and the single-expert model, which is indeed the Topic model, performs the worst. Besides, simply increasing the number of experts does not imply a gain in performance. For example, the performance of the model \textit{n256\_d128} ($\sim$27.2M parameters) is worse than the best-performing model \textit{n8\_d256} ($\sim$17.9M parameters). 

\subsubsection{Topic-wise Result Comparison}
To examine the performance of our method on each novel activity, we report the topic-wise comparison with the Base model in \autoref{table:pertopic}. The TAMoE model outperforms the Base model on most of the activities (12 out of 15), of which some activities are improved by a remarkable margin, e.g., \emph{arm wresting}, \emph{braiding hair}, \emph{gargling mouthwash}, and \emph{sharpening knives}. Meanwhile, we showcase the top-4 related words from the external corpus for each topic according to their TF-IDF weights to provide a better illustration of our topic embeddings.   

\subsubsection{Qualitative Comparison}
\autoref{fig:demo} showcases two qualitative examples on the unseen test set. In the first video about ``\emph{painting fence}'', the Base model has no linguistic knowledge of the concept ``\emph{fence}'', while our TAMoE model successfully recognizes it and produces a more pertinent description. In the second example about ``\emph{grooming dog}'', the Base model fails to recognize the actual action though already knowing the objects, while our model generates a more accurate description of the video.   

\section{Discussion}
In this paper, we formally define the task of zero-shot video captioning and set up a common setting for evaluation. In order to accurately describe videos of unseen activities, we seek solutions based on what and how to utilize and transfer. 
Note that one assumption of zero-shot video captioning is that the activity category can be either provided or predicted. Even so, it is still valuable because caption annotations for videos are much more expensive to get compared with activity labels. But combining zero-shot activity recognition and zero-shot video captioning is a promising direction towards more advanced approaches for transfer learning, which we leave for future study.

\section{Acknowledgement}
We thank Adobe Research for supporting our language
and vision research. We also thank the anonymous reviewers for their helpful feedbacks and Yijun Xiao for cleaning the data. 

\bibliography{aaai}

\begin{thebibliography}{}

\bibitem[\protect\citeauthoryear{Ahmed, Baig, and
  Torresani}{2016}]{ahmed2016network}
Ahmed, K.; Baig, M.~H.; and Torresani, L.
\newblock 2016.
\newblock Network of experts for large-scale image categorization.
\newblock In {\em ECCV}.

\bibitem[\protect\citeauthoryear{Anderson \bgroup et al\mbox.\egroup
  }{2017}]{anderson-EtAl:2017:EMNLP2017}
Anderson, P.; Fernando, B.; Johnson, M.; and Gould, S.
\newblock 2017.
\newblock Guided open vocabulary image captioning with constrained beam search.
\newblock In {\em EMNLP}.

\bibitem[\protect\citeauthoryear{Anne~Hendricks \bgroup et al\mbox.\egroup
  }{2016}]{anne2016deep}
Anne~Hendricks, L.; Venugopalan, S.; Rohrbach, M.; Mooney, R.; Saenko, K.;
  Darrell, T.; Mao, J.; Huang, J.; Toshev, A.; Camburu, O.; et~al.
\newblock 2016.
\newblock Deep compositional captioning: Describing novel object categories
  without paired training data.
\newblock In {\em CVPR}.

\bibitem[\protect\citeauthoryear{Bahdanau, Cho, and
  Bengio}{2015}]{bahdanau2014neural}
Bahdanau, D.; Cho, K.; and Bengio, Y.
\newblock 2015.
\newblock Neural machine translation by jointly learning to align and
  translate.
\newblock In {\em ICLR}.

\bibitem[\protect\citeauthoryear{Bengio \bgroup et al\mbox.\egroup
  }{2015}]{bengio2015scheduled}
Bengio, S.; Vinyals, O.; Jaitly, N.; and Shazeer, N.
\newblock 2015.
\newblock Scheduled sampling for sequence prediction with recurrent neural
  networks.
\newblock In {\em NIPS}.

\bibitem[\protect\citeauthoryear{Carreira and
  Zisserman}{2017}]{carreira2017quo}
Carreira, J., and Zisserman, A.
\newblock 2017.
\newblock Quo vadis, action recognition? a new model and the kinetics dataset.
\newblock In {\em CVPR}.

\bibitem[\protect\citeauthoryear{Collobert, Bengio, and
  Bengio}{2002}]{collobert2002parallel}
Collobert, R.; Bengio, S.; and Bengio, Y.
\newblock 2002.
\newblock A parallel mixture of svms for very large scale problems.
\newblock In {\em NIPS}.

\bibitem[\protect\citeauthoryear{Collobert, Bengio, and
  Bengio}{2003}]{collobert2003scaling}
Collobert, R.; Bengio, Y.; and Bengio, S.
\newblock 2003.
\newblock Scaling large learning problems with hard parallel mixtures.
\newblock {\em International Journal of pattern recognition and artificial
  intelligence} 17(03):349--365.

\bibitem[\protect\citeauthoryear{Fabian Caba~Heilbron and
  Niebles}{2015}]{caba2015activitynet}
Fabian Caba~Heilbron, Victor~Escorcia, B.~G., and Niebles, J.~C.
\newblock 2015.
\newblock Activitynet: A large-scale video benchmark for human activity
  understanding.
\newblock In {\em CVPR}.

\bibitem[\protect\citeauthoryear{Gal and
  Ghahramani}{2016}]{gal2016theoretically}
Gal, Y., and Ghahramani, Z.
\newblock 2016.
\newblock A theoretically grounded application of dropout in recurrent neural
  networks.
\newblock In {\em NIPS}.

\bibitem[\protect\citeauthoryear{Gan \bgroup et al\mbox.\egroup
  }{2015}]{gan2015exploring}
Gan, C.; Liu, M.; Yang, Y.; Zhuang, Y.; and Hauptmann, A.~G.
\newblock 2015.
\newblock Exploring semantic interclass relationships (sir) for zero-shot
  action recognition.
\newblock In {\em AAAI}.

\bibitem[\protect\citeauthoryear{Gan \bgroup et al\mbox.\egroup
  }{2016}]{gan2016concepts}
Gan, C.; Lin, M.; Yang, Y.; de~Melo, G.; and Hauptmann, A.~G.
\newblock 2016.
\newblock Concepts not alone: Exploring pairwise relationships for zero-shot
  video activity recognition.
\newblock In {\em AAAI}.

\bibitem[\protect\citeauthoryear{Gan \bgroup et al\mbox.\egroup
  }{2017}]{gan2016semantic}
Gan, Z.; Gan, C.; He, X.; Pu, Y.; Tran, K.; Gao, J.; Carin, L.; and Deng, L.
\newblock 2017.
\newblock Semantic compositional networks for visual captioning.
\newblock In {\em CVPR}.

\bibitem[\protect\citeauthoryear{Gu \bgroup et al\mbox.\egroup
  }{2018}]{gu2018naacl}
Gu, J.; Hassan, H.; Devlin, J.; and Li, V.~O.
\newblock 2018.
\newblock Universal neural machine translation for extremely low resource
  languages.
\newblock In {\em NAACL HLT}.

\bibitem[\protect\citeauthoryear{Jacobs \bgroup et al\mbox.\egroup
  }{1991}]{jacobs1991adaptive}
Jacobs, R.~A.; Jordan, M.~I.; Nowlan, S.~J.; and Hinton, G.~E.
\newblock 1991.
\newblock Adaptive mixtures of local experts.
\newblock {\em Neural computation} 3(1):79--87.

\bibitem[\protect\citeauthoryear{Jordan and
  Jacobs}{1994}]{jordan1994hierarchical}
Jordan, M.~I., and Jacobs, R.~A.
\newblock 1994.
\newblock Hierarchical mixtures of experts and the em algorithm.
\newblock {\em Neural computation} 6(2):181--214.

\bibitem[\protect\citeauthoryear{Kay \bgroup et al\mbox.\egroup
  }{2017}]{kay2017kinetics}
Kay, W.; Carreira, J.; Simonyan, K.; Zhang, B.; Hillier, C.; Vijayanarasimhan,
  S.; Viola, F.; Green, T.; Back, T.; Natsev, P.; et~al.
\newblock 2017.
\newblock The kinetics human action video dataset.
\newblock {\em arXiv preprint arXiv:1705.06950}.

\bibitem[\protect\citeauthoryear{Krishna \bgroup et al\mbox.\egroup
  }{2017}]{krishna2017dense}
Krishna, R.; Hata, K.; Ren, F.; Fei-Fei, L.; and Niebles, J.~C.
\newblock 2017.
\newblock Dense-captioning events in videos.
\newblock In {\em ICCV}.

\bibitem[\protect\citeauthoryear{Lu \bgroup et al\mbox.\egroup
  }{2018}]{lu2018neural}
Lu, J.; Yang, J.; Batra, D.; and Parikh, D.
\newblock 2018.
\newblock Neural baby talk.
\newblock In {\em CVPR}.

\bibitem[\protect\citeauthoryear{Merity, Keskar, and
  Socher}{2018}]{merity2017regularizing}
Merity, S.; Keskar, N.~S.; and Socher, R.
\newblock 2018.
\newblock Regularizing and optimizing lstm language models.
\newblock In {\em ICLR}.

\bibitem[\protect\citeauthoryear{Mikolov \bgroup et al\mbox.\egroup
  }{2018}]{mikolov2018advances}
Mikolov, T.; Grave, E.; Bojanowski, P.; Puhrsch, C.; and Joulin, A.
\newblock 2018.
\newblock Advances in pre-training distributed word representations.
\newblock In {\em LREC}.

\bibitem[\protect\citeauthoryear{Pan \bgroup et al\mbox.\egroup
  }{2016}]{pan2016hierarchical}
Pan, P.; Xu, Z.; Yang, Y.; Wu, F.; and Zhuang, Y.
\newblock 2016.
\newblock Hierarchical recurrent neural encoder for video representation with
  application to captioning.
\newblock In {\em CVPR}.

\bibitem[\protect\citeauthoryear{Pasunuru and
  Bansal}{2017}]{pasunuru-bansal:2017:Long}
Pasunuru, R., and Bansal, M.
\newblock 2017.
\newblock Multi-task video captioning with video and entailment generation.
\newblock In {\em ACL}.

\bibitem[\protect\citeauthoryear{Rohrbach \bgroup et al\mbox.\egroup
  }{2014}]{rohrbach14gcpr}
Rohrbach, A.; Rohrbach, M.; Qiu, W.; Friedrich, A.; Pinkal, M.; and Schiele, B.
\newblock 2014.
\newblock Coherent multi-sentence video description with variable level of
  detail.
\newblock In {\em GCPR}.

\bibitem[\protect\citeauthoryear{Shazeer \bgroup et al\mbox.\egroup
  }{2017}]{shazeer2017outrageously}
Shazeer, N.; Mirhoseini, A.; Maziarz, K.; Davis, A.; Le, Q.; Hinton, G.; and
  Dean, J.
\newblock 2017.
\newblock Outrageously large neural networks: The sparsely-gated
  mixture-of-experts layer.
\newblock In {\em ICLR}.

\bibitem[\protect\citeauthoryear{Shen \bgroup et al\mbox.\egroup
  }{2017}]{Shen_2017_CVPR}
Shen, Z.; Li, J.; Su, Z.; Li, M.; Chen, Y.; Jiang, Y.-G.; and Xue, X.
\newblock 2017.
\newblock Weakly supervised dense video captioning.
\newblock In {\em CVPR}.

\bibitem[\protect\citeauthoryear{Thomason \bgroup et al\mbox.\egroup
  }{2014}]{thomason2014integrating}
Thomason, J.; Venugopalan, S.; Guadarrama, S.; Saenko, K.; and Mooney, R.
\newblock 2014.
\newblock Integrating language and vision to generate natural language
  descriptions of videos in the wild.
\newblock In {\em COLING}.

\bibitem[\protect\citeauthoryear{Tresp}{2001}]{tresp2001mixtures}
Tresp, V.
\newblock 2001.
\newblock Mixtures of gaussian processes.
\newblock In {\em NIPS}.

\bibitem[\protect\citeauthoryear{Venugopalan \bgroup et al\mbox.\egroup
  }{2015}]{venugopalan2015sequence}
Venugopalan, S.; Rohrbach, M.; Donahue, J.; Mooney, R.; Darrell, T.; and
  Saenko, K.
\newblock 2015.
\newblock Sequence to sequence-video to text.
\newblock In {\em ICCV}.

\bibitem[\protect\citeauthoryear{Venugopalan \bgroup et al\mbox.\egroup
  }{2016}]{venugopalan2016improving}
Venugopalan, S.; Hendricks, L.~A.; Mooney, R.; and Saenko, K.
\newblock 2016.
\newblock Improving lstm-based video description with linguistic knowledge
  mined from text.
\newblock In {\em EMNLP}.

\bibitem[\protect\citeauthoryear{Venugopalan \bgroup et al\mbox.\egroup
  }{2017}]{venugopalan2016captioning}
Venugopalan, S.; Hendricks, L.~A.; Rohrbach, M.; Mooney, R.; Darrell, T.; and
  Saenko, K.
\newblock 2017.
\newblock Captioning images with diverse objects.
\newblock In {\em CVPR}.

\bibitem[\protect\citeauthoryear{Wang \bgroup et al\mbox.\egroup
  }{2016}]{wang2016temporal}
Wang, L.; Xiong, Y.; Wang, Z.; Qiao, Y.; Lin, D.; Tang, X.; and Van~Gool, L.
\newblock 2016.
\newblock Temporal segment networks: Towards good practices for deep action
  recognition.
\newblock In {\em ECCV}.

\bibitem[\protect\citeauthoryear{Wang \bgroup et al\mbox.\egroup
  }{2018a}]{wang2018AREL}
Wang, X.; Chen, W.; Wang, Y.-F.; and Wang, W.~Y.
\newblock 2018a.
\newblock No metrics are perfect: Adversarial reward learning for visual
  storytelling.
\newblock In {\em ACL}.

\bibitem[\protect\citeauthoryear{Wang \bgroup et al\mbox.\egroup
  }{2018b}]{wang2018video}
Wang, X.; Chen, W.; Wu, J.; Wang, Y.-F.; and Wang, W.~Y.
\newblock 2018b.
\newblock Video captioning via hierarchical reinforcement learning.
\newblock In {\em CVPR}.

\bibitem[\protect\citeauthoryear{Wang \bgroup et al\mbox.\egroup
  }{2018c}]{wang2018deep}
Wang, X.; Yu, F.; Wang, R.; Ma, Y.-A.; Mirhoseini, A.; Darrell, T.; and
  Gonzalez, J.~E.
\newblock 2018c.
\newblock Deep mixture of experts via shallow embedding.
\newblock {\em arXiv preprint arXiv:1806.01531}.

\bibitem[\protect\citeauthoryear{Wang, Wang, and Wang}{2018}]{wang2018watch}
Wang, X.; Wang, Y.-F.; and Wang, W.~Y.
\newblock 2018.
\newblock Watch, listen, and describe: Globally and locally aligned cross-modal
  attentions for video captioning.
\newblock {\em {NAACL} {HLT}}.

\bibitem[\protect\citeauthoryear{Wu \bgroup et al\mbox.\egroup
  }{2018}]{wu2018decoupled}
Wu, Y.; Zhu, L.; Jiang, L.; and Yang, Y.
\newblock 2018.
\newblock Decoupled novel object captioner.
\newblock In {\em ACM MM}.

\bibitem[\protect\citeauthoryear{Xu \bgroup et al\mbox.\egroup
  }{2016}]{xu2016msr}
Xu, J.; Mei, T.; Yao, T.; and Rui, Y.
\newblock 2016.
\newblock Msr-vtt: A large video description dataset for bridging video and
  language.
\newblock In {\em CVPR}.

\bibitem[\protect\citeauthoryear{Yang \bgroup et al\mbox.\egroup
  }{2018}]{yang2018breaking}
Yang, Z.; Dai, Z.; Salakhutdinov, R.; and Cohen, W.~W.
\newblock 2018.
\newblock Breaking the softmax bottleneck: A high-rank {RNN} language model.
\newblock In {\em ICLR}.

\bibitem[\protect\citeauthoryear{Yao \bgroup et al\mbox.\egroup
  }{2015}]{yao2015describing}
Yao, L.; Torabi, A.; Cho, K.; Ballas, N.; Pal, C.; Larochelle, H.; and
  Courville, A.
\newblock 2015.
\newblock Describing videos by exploiting temporal structure.
\newblock In {\em ICCV}.

\bibitem[\protect\citeauthoryear{Yu \bgroup et al\mbox.\egroup
  }{2016}]{yu2016video}
Yu, H.; Wang, J.; Huang, Z.; Yang, Y.; and Xu, W.
\newblock 2016.
\newblock Video paragraph captioning using hierarchical recurrent neural
  networks.
\newblock In {\em CVPR}.

\bibitem[\protect\citeauthoryear{Zeiler}{2012}]{zeiler2012adadelta}
Zeiler, M.~D.
\newblock 2012.
\newblock Adadelta: an adaptive learning rate method.
\newblock {\em arXiv preprint arXiv:1212.5701}.

\bibitem[\protect\citeauthoryear{Zellers and Choi}{2017}]{zellers2017zero}
Zellers, R., and Choi, Y.
\newblock 2017.
\newblock Zero-shot activity recognition with verb attribute induction.
\newblock In {\em EMNLP}.

\bibitem[\protect\citeauthoryear{Zhang \bgroup et al\mbox.\egroup
  }{2018}]{zhang2018bmvc}
Zhang, D.; Dai, X.; Wang, X.; and Wang, Y.-F.
\newblock 2018.
\newblock S3d: Single shot multi-span detector via fully 3d convolutional
  network.
\newblock In {\em BMVC}.

\end{thebibliography}
\bibliographystyle{aaai}

\end{document}